\documentclass[conference]{IEEEtran}
\IEEEoverridecommandlockouts
\usepackage{cite}
\usepackage{amsmath,amssymb,amsfonts}
\usepackage{algorithmic}
\usepackage{graphicx}
\usepackage{textcomp}
\usepackage{xcolor}
\def\BibTeX{{\rm B\kern-.05em{\sc i\kern-.025em b}\kern-.08em
    T\kern-.1667em\lower.7ex\hbox{E}\kern-.125emX}}
    
\usepackage{mathtools}
\usepackage{cuted}
    
\usepackage{listings}
\usepackage{xcolor}

\colorlet{punct}{red!60!black}
\definecolor{background}{HTML}{FFFFFF}
\definecolor{delim}{RGB}{20,105,176}
\definecolor{key}{RGB}{20,25,175}
\definecolor{bool}{RGB}{20,85,75}
\colorlet{numb}{magenta!60!black}

\lstdefinelanguage{json}{
    basicstyle=\small\ttfamily,
    numbers=left,
    numberstyle=\scriptsize,
    stepnumber=1,
    numbersep=8pt,
    showstringspaces=false,
    breaklines=true,
    backgroundcolor=\color{background},
    literate=
     *{0}{{{\color{numb}0}}}{1}
      {1}{{{\color{numb}1}}}{1}
      {2}{{{\color{numb}2}}}{1}
      {3}{{{\color{numb}3}}}{1}
      {4}{{{\color{numb}4}}}{1}
      {5}{{{\color{numb}5}}}{1}
      {6}{{{\color{numb}6}}}{1}
      {7}{{{\color{numb}7}}}{1}
      {8}{{{\color{numb}8}}}{1}
      {9}{{{\color{numb}9}}}{1}
      {QH}{{{\color{numb}QH}}}{2}
      {q-}{{{\color{numb}q-}}}{2}
      {/}{{{\color{numb}/}}}{1}
      {question\_id}{{{\color{key}question\_id}}}{11}
      {text}{{{\color{key}text}}}{4}
      {relevant\_article}{{{\color{key}relevant\_article}}}{16}
      {label}{{{\color{key}label}}}{5}
      {law\_id}{{{\color{key}law\_id}}}{6}
      {article\_id}{{{\color{key}article\_id}}}{10}
      {true}{{{\color{bool}true}}}{4}
      {false}{{{\color{bool}false}}}{5}
      {:}{{{\color{punct}{:}}}}{1}
      {,}{{{\color{punct}{,}}}}{1}
      {\{}{{{\color{delim}{\{}}}}{1}
      {\}}{{{\color{delim}{\}}}}}{1}
      {[}{{{\color{delim}{[}}}}{1}
      {]}{{{\color{delim}{]}}}}{1},
}
\begin{document}

\title{A Summary of the ALQAC 2021 Competition
}

\author{\IEEEauthorblockN{
    Nguyen Ha Thanh\textsuperscript{1,*}\thanks{\textsuperscript{*} Corresponding: nguyenhathanh@jaist.ac.jp.}, 
    Bui Minh Quan\textsuperscript{1},
    Chau Nguyen\textsuperscript{1},
    Tung Le\textsuperscript{1}, 
    Nguyen Minh Phuong\textsuperscript{1},\\
    Dang Tran Binh\textsuperscript{1},
    Vuong Thi Hai Yen\textsuperscript{2},
    Teeradaj Racharak\textsuperscript{1},
    Nguyen Le Minh\textsuperscript{1},
    Tran Duc Vu\textsuperscript{3},\\
    Phan Viet Anh\textsuperscript{4}, 
    Nguyen Truong Son\textsuperscript{5},
    Huy Tien Nguyen\textsuperscript{5},
    Bhumindr Butr-indr\textsuperscript{6}, \\
    Peerapon Vateekul\textsuperscript{7}, 
    Prachya Boonkwan\textsuperscript{8}, 
}
\IEEEauthorblockA{
\textit{\textsuperscript{1}Japan Advanced Institute of Science and Technology}\\
Ishikawa, Japan\\
\textit{\textsuperscript{2}VNU University of Engineering and Technology}\\
Hanoi, Vietnam\\
\textit{\textsuperscript{3}The Institute of Statistical Mathematics (ISM), Japan}\\
Tokyo, Japan\\
\textit{\textsuperscript{4}Le Quy Don Technical University (LQDTU), Vietnam}\\
Hanoi, Vietnam\\
\textit{\textsuperscript{5}Ho Chi Minh University of Science (VNU-HCMUS), Vietnam}\\
Ho Chi Minh City, Vietnam\\
\textit{\textsuperscript{6}Faculty of Law, Thammasat University (TU), Thailand}\\
Bangkok, Thailand\\
\textit{\textsuperscript{7}Chulalongkorn University (CU), Thailand}\\
Bangkok, Thailand\\
\textit{\textsuperscript{8}National Electronics and Computer Technology Center (NECTEC), Thailand}\\
Pathumthani, Thailand
}}
\maketitle

\begin{abstract}
We summarize the evaluation of the first Automated Legal Question Answering Competition (ALQAC 2021). 
The competition this year contains three tasks, which aims at processing the statute law document, which are Legal Text Information Retrieval (Task~1), Legal Text Entailment Prediction (Task 2), and Legal Text Question Answering (Task 3).
The final goal of these tasks is to build a system that can automatically determine whether a particular statement is lawful. 
There is no limit to the approaches of the participating teams.
This year, there are 5 teams participating in Task 1, 6 teams participating in Task 2, and 5 teams participating in Task 3.
There are in total 36 runs submitted to the organizer.
In this paper, we summarize each team’s approaches, official results, and some discussion about the competition.
Only results of the teams who successfully submit their approach description paper are reported in this paper.

\end{abstract}

\begin{IEEEkeywords}
ALQAC 2021, summary, legal processing, competition
\end{IEEEkeywords}

\section{Introduction}
The achievements of natural language processing in recent years are remarkable. 
Their applications are present in almost every industry.
Automated Legal Question Answering Competition (ALQAC 2021) is held with the goal of building a research community in legal text processing and collecting the idea of applying the most advanced techniques to solve law-related problems.
ALQAC 2021 is the first time this competition take place, co-located with KSE, International Conference on Knowledge and Systems Engineering Conference.
Knowledge and system engineering are the two factors that mostly affect the performance of the system in legal text processing.
Inspired by the Competition on Legal Information Extraction/Entailment (COLIEE) \cite{juliano2021summary} for English and Japanese, ALQAC is designed for low-resource language with its own language challenges.

In ALQAC 2021, there are three tasks in statute law processing (information retrieval, entailment, and question answering). 
The competition's data is prepared in Vietnamese and Thai language.
However, this year, all of the participants only choose to work on the Vietnamese dataset.
Task~1 is a statute law retrieval task, given a legal query, the systems need to retrieve the most relevant articles. 
We evaluate the systems using the F2 score metric to balance between precision and recall with precision is weighted higher.
In Task~2, the systems can use the retrieval results from Task 1 to make a prediction of whether the articles entail the given query or not, and from that, answer if the statement in the query is lawful or non-lawful.
In Task~3, the systems need to directly predict the lawfulness of the query without using the relevant articles.
The evaluation metric for Task 2 and Task 3 is accuracy.

The rest of the paper is organized as follows: Section \ref{sec:dataset} describes the dataset and evaluation metrics, Sections \ref{sec:task1}, \ref{sec:task2}, \ref{sec:task3} describe each task, presenting their definitions, list of approaches submitted by the participants, and results attained. Section \ref{sec:remark} presents some final remarks.

\section{Dataset}
\label{sec:dataset}

Legal data used in ALQAC 2021 are Vietnamese and Thai legal documents.
Based on the legal text, we pose questions that can be answered using the relevant articles alone without the need for additional sources such as legal theory or supporting evidence.
The questions as well as the relevant articles are verified by legal experts.
Finally, the dataset is formatted to make it most convenient for data processing and result evaluation.

The dataset file formats are shown via examples as follows.

\begin{enumerate}
    \item \textbf{Legal Articles:} Details about each article are in the following format:
\vspace{2mm}
\begin{lstlisting}[language=json,numbers=none]
[
    {
      "id": "45/2019/QH14",
      "articles": [
            {
                "text": "The content of legal article",
                "id": "1"
            }
        ]
    }
]
\end{lstlisting}  
    \item \textbf{Annotation Samples:} Details about each sample are in the following format:
\vspace{2mm}
\begin{lstlisting}[language=json,numbers=none]
[
    {
        "question_id": "q-1",
        "text": "The content of question or statement",
        "label": true,
        "relevant_articles": [
            {
                "law_id": "45/2019/QH14",
                "article_id": "1"
            }
        ]
    }
]
\end{lstlisting}
\end{enumerate}

\begin{strip}
\begin{align}
    Precision_i &=   \frac{\textrm{the number of correctly retrieved articles of query $i^{th}$}}{\textrm{the number of retrieved articles of query $i^{th}$}} \label{eq:precision_i}\\
    Recall_i &=   \frac{\textrm{the number of correctly retrieved articles of query $i^{th}$}}{\textrm{the number of relevant articles) of query $i^{th}$}} \label{eq:recall_i}\\
    F2_i &=    \frac{(5 \times Precision_i \times Recall_i)}{(4\times Precision_i + Recall_i)} \label{eq:f2_i}\\
    Marcro-F2 &= \mathrm{Average\ of}(F2_i) \label{eq:f2}\\
    Accuracy &= \frac{\textrm{(the number of queries which were correctly confirmed as true or false)}}{\textrm{(the number of all queries)}} \label{eq:acc}
\end{align}
\end{strip}

The evaluation metrics for Task 1 are precision, recall, and Macro-F2 score as in Equations \ref{eq:precision_i}-\ref{eq:f2}. Accuracy is used to evaluate results in Task 2 and Task 3 with respect to whether the yes/no question was correctly confirmed in Equation \ref{eq:acc}.

\vspace{.5cm}
In ALQAC 2021, the method used to calculate the final F2 score of all queries is macro-average (evaluation measure is calculated for each query and their average is used as the final evaluation measure) instead of micro-average (evaluation measure is calculated using results of all queries).

\vspace{1cm}
\section{Task 1 - Legal Information Retrieval}
\label{sec:task1}
\subsection{Task Description}
Task 1’s goal is to return articles that are related to a giving statement. An article is considered ``relevant'' to a statement \textit{iff} the statement rightness can be entailed (as Yes/No) by the article. This task requires the retrieval of all the articles that are relevant to a query.

Specifically, the input samples consist of:
\begin{enumerate}
    \item \textbf{Legal Articles:} whose format is the same as Legal Articles described in Section \ref{sec:dataset}.
    \vspace{2mm}
    \item \textbf{Question:} whose format is in JSON as follows: 
    
    \begin{lstlisting}[language=json,numbers=none]
[
    {
        "question_id": "q-1",
        "text": "The content of question or statement"
    }
]
    \end{lstlisting}
\end{enumerate}

The system should retrieve all the relevant articles as follows:

\vspace{2mm}
\begin{lstlisting}[language=json,numbers=none]
[
    {
        "question_id": "q-193",
        "relevant_articles": [
            {
                "law_id": "100/2015/QH13",
                "article_id": "177"
            }
        ]
    },
    ...
]
\end{lstlisting}
\vspace{2mm}
\noindent In which, ``relevant\_articles'' are the list of all relevant articles of the questions/statements.

\vspace{1cm}
\subsection{Approaches}
There are in total 10 validated runs submitted in this task.
\begin{enumerate}
    \item \textbf{AimeLaw (3 runs)} \cite{ngo2021aimelaw} propose approaches of combining scores of BM25 with  Domain Invariant Supporting Model and Deep CNN Supporting Model using a weighted sum function.
    \item \textbf{Aleph (1 run)} \cite{truong2021apply} build an article ranking model by finetuning their own pretrained model VNLawBERT \cite{chau2020vnlawbert} with a binary classification problem. The authors make negative samples by choosing the closest candidate with the gold samples.
    \item \textbf{Kodiak (3 runs)} \cite{do2021kodiak} use the lexical matching methods along with the semantic search models with the goal of reaching a balance between the lexical and semantic features.
    \item \textbf{Dline (3 runs)} \cite{luu2021vietnamese} preprocess the data to extract lexical features and word embedding features and feeds them into an XGBoost classifier to predict whether a pair of a query and an article is relevant or not.
\end{enumerate}

\vspace{1cm}
\subsection{Results}
The final results in this task can be seen in Table~\ref{tab:task1_res}.
The performance of the teams is generally not too disparate.
All teams surpass the TF-IDF baseline which is $0.6392$ F2 Score.
With only one run, \textbf{Aleph} surprisingly achieve state-of-the-art performance with $0.8807$ F2 Score, followed by two runs of \textbf{AimeLaw} with $0.8061$ F2 Score.

\begin{table}
\centering
\caption{Results on Task~1: The bold result indicates the first prize, the underlined result indicates the second prize.}
\label{tab:task1_res}
\begin{tabular}{llr}
\hline
\textbf{Team} & \textbf{Run ID}                     & \textbf{F2 Score}     \\ \hline
AimeLaw     & run1\_bm25                                    & 0.7842               \\
AimeLaw     & run2\_bm25\_bert\_cnn\_09\_threshold          & \underline{0.8061}             \\
AimeLaw     & run3\_bm25\_bert\_domain\_09\_threshold       & \underline{0.8061}               \\
Aleph       & result\_task\_1                               & \textbf{0.8807}    \\
Kodiak      & Task\_1\_run\_1                                  & 0.7855    \\
Kodiak      & Task\_1\_run\_2                                  & 0.7919    \\
Kodiak      & Task\_1\_run\_3                                  & 0.7955    \\
Dline       & result\_task\_1\_ver1                            & 0.7766    \\
Dline       & result\_task\_1\_ver2                            & 0.7776    \\
Dline       & result\_task\_1\_ver3                            & 0.7936    \\
\hline
\end{tabular}

\end{table}
\vspace{1cm}
\section{Task 2 - Legal Textual Entailment}
\label{sec:task2}
\subsection{Task Description}
Task 2's goal is to construct Yes/No question answering systems for legal queries, by entailment from the relevant articles. Based on the content of legal articles, the system should answer whether the statement is true or false.

Specifically, the input samples consist of the pair of question/statement and relevant articles ($>=1$) as follows:
\vspace{2mm}
\begin{lstlisting}[language=json,numbers=none]
[
    {
        "question_id": "q-1",
        "text": "The content of question or statement",
        "relevant_articles": [
            {
                "law_id": "45/2019/QH14",
                "article_id": "1"
            }
        ]
    }
]
\end{lstlisting}

The system should answer whether the statement is true or false via ``label'' in JSON format as follows:
\vspace{2mm}
\begin{lstlisting}[language=json,numbers=none]
[
    {
        "question_id": "q-193",
        "label": false
    },
    ...
]
\end{lstlisting}

The evaluation measure is accuracy, with respect to whether the yes/no question was correctly confirmed as in Equation \ref{eq:acc}.

\subsection{Approaches}
The participants submit 11 valid runs for Task 2 (legal textual entailment).
\begin{enumerate}
    \item \textbf{AimeLaw (3 runs)} \cite{ngo2021aimelaw} propose to fine-tune a BERT model on the sentence pair data created by applying their data augmentation and text matching technique on the annotated data.
    \item \textbf{Aleph (2 runs)} \cite{truong2021apply} also utilize a BERT model to be fine-tuned on sentence pair data. However, instead of augmenting data based on the annotated data, they collect external data from law-related websites.
    \item \textbf{Kodiak (3 runs)} \cite{do2021kodiak} propose to fine-tune BERT models (Multilingual-BERT and PhoBERT \cite{nguyen2020phobert}) on the annotated data, with/without an asymmetric truncation technique (to address the limitation of BERT models when dealing with long sentences).
    \item \textbf{Dline (3 runs)} \cite{luu2021vietnamese} propose two directions: (i) utilize classifiers (Random Forest and SVM) on TF-IDF features, (ii) fine-tune BERT models on the sequence classification task.
\end{enumerate}

\subsection{Results}
We use the proportion of the majority class as a baseline with an accuracy of $0.5455$.
Table~\ref{tab:task2_res} provides the results in accuracy of participants' models for Task 2. 
Most of the teams' performance surpasses the baseline with a significant gap.
The highest accuracy, $0.6989$, is achieved by \textbf{AimeLaw} and \textbf{Aleph}, and the second-highest accuracy belongs to \textbf{Kodiak} with $0.6818$. In this task, all participants make use of \textit{BERT} pretrained models. It indicates the robust ability of large-scale pretrained models in language understanding, and their adaptation ability to the legal domain.

\begin{table}
\centering
\caption{Results on Task~2: The bold result indicates the first prize, the underlined result indicates the second prize.}
\label{tab:task2_res}
\begin{tabular}{llr}
\hline
\textbf{Team} & \textbf{Run ID}                     & \textbf{Accuracy}     \\ \hline
AimeLaw     & aimelaw\_predictions\_1\_                     & \textbf{0.6989}               \\
AimeLaw     & aimelaw\_predictions\_2\_                     & 0.6761             \\
AimeLaw     & aimelaw\_predictions\_3\_                     & 0.4318               \\
Aleph       & aleph\_single                             & 0.5398    \\
Aleph       & aleph\_pair                               & \textbf{0.6989}    \\
Kodiak      & Task\_2\_run\_1                                  & 0.6364    \\
Kodiak      & Task\_2\_run\_2                                  & \underline{0.6818}    \\
Kodiak      & Task\_2\_run\_3                                  & 0.5568    \\
Dline       & result\_task\_2\_ver1                            & 0.4034    \\
Dline       & result\_task\_2\_ver2                            & 0.4148    \\
Dline       & result\_task\_2\_ver3                            & 0.4148    \\
\hline
\end{tabular}

\end{table}

\section{Task 3 - Legal Question Answering}
\label{sec:task3}
\subsection{Task Description}
Task 3’s goal is to construct Yes/No question answering systems for legal queries.

Given a legal statement or legal question, the task is to answer “Yes” or “No”, in other words, to determine whether it is true or false. This question answering could be a concatenation of Task 1 and Task 2, but not necessarily so, e.g. using any knowledge source other than the results of Task 2.

Specifically, the input samples consist of the question/statement as follows:
\vspace{2mm}
\begin{lstlisting}[language=json,numbers=none]
[
    {
        "question_id": "q-1",
        "text": "The content of question or statement"
    }
]
\end{lstlisting}

The system is not provided with the relevant articles. As a result, the teams can use their own results in Task 1 or another source of knowledge to automated answer whether the question/statement is true or false via ``label'' in JSON format as same as in Task 2:
\vspace{2mm}
\begin{lstlisting}[language=json,numbers=none]
[
    {
        "question_id": "q-194",
        "label": true
    },
    ...
]
\end{lstlisting}

\subsection{Approaches}
In this task, we receive 9 different valid runs from 4 participants. The summary information about each team's approach is as below:
\begin{enumerate}
    \item \textbf{AimeLaw (3 runs)} \cite{ngo2021aimelaw} leverage their system in Task 1. They use Task 1 pretrained BERT \cite{devlin2018bert} model to classify whether the giving query is Yes or No. By using a Text Matching technique, they find the most relevant clause for a giving query within the articles, then use this clause and the giving query as the input of BERT classifier.
    \item \textbf{Aleph (2 runs)} \cite{truong2021apply} use a pretrained model as the classifier with the training data prepared as in the two following methods:
    \begin{itemize}
        \item Utilizing Task 1 and Task 2 outputs for generating training data.
        \item Using provided training data and generating more positive samples by filtering clauses in provided articles.
    \end{itemize}
    \item \textbf{Kodiak (3 runs)} \cite{do2021kodiak} use output of Task 1 and Task 2 to train a sentence classifier model based on BERT \cite{devlin2018bert}, and use this model to make predictions on Task 3.

    \item \textbf{Dline (3 runs)} \cite{luu2021vietnamese} suggest an idea of separating an article into meaningful sentences. Based on this segmentation, Dline use a pair of a single sentence and a giving query as the sample to fine-tune on PhoBert.
\end{enumerate}

\subsection{Results}
The baseline for Task 3 is also 0.5455 in accuracy, we can see in Table \ref{tab:task3_res}, 7 of 11 runs are better than the baseline. 
The common idea of participated teams for Task 3 is to use outputs of Task 1 and Task 2 for training a classifier. 
Therefore, \textbf{Aleph} achieves first place on Task 3 with \textit{0.7102} accuracy due to impressive results on Task 1 \textit{0.8807} on F2 score, and Task 2 \textit{0.6989}.

\begin{table}
\centering
\caption{Results on Task~3: The bold result indicates the first prize, the underlined result indicates the second prize.}
\label{tab:task3_res}
\begin{tabular}{llr}
\hline
\textbf{Team} & \textbf{Run ID}                     & \textbf{Accuracy}     \\ \hline
AimeLaw     & run1\_bm25+bert                               & \underline{0.6477}               \\
AimeLaw     & run2\_bm25\_bert\_domain                      & 0.6136             \\
AimeLaw     & run3\_bm25\_bert\_cnn                         & 0.6136               \\
Aleph       & result\_task\_3\_pair                         & \textbf{0.7102}    \\
Aleph       & result\_task\_3\_single                       & 0.5398    \\
Kodiak      & Task\_3\_run\_1                               & 0.5739    \\
Kodiak      & Task\_3\_run\_2                               & 0.5682    \\
Kodiak      & Task\_3\_run\_3                               & 0.6250    \\
Dline       & result\_task\_3\_ver1                         & 0.5114    \\
Dline       & result\_task\_3\_ver2                         & 0.5057    \\
Dline       & result\_task\_3\_ver3                         & 0.5170    \\
\hline
\end{tabular}

\end{table}

\vspace{.5cm}
\section{Final Remarks}
\label{sec:remark}
We have summarized the results of the ALQUAC 2021 competition. 
Task 1 is about retrieving articles for the purpose of verifying the lawfulness of a given legal question. 
Task 2 and Task 3 to confirm whether the legal question is lawful or not with and without the relevant articles. 
We received results from 5 teams for Task 1 (a total of 11 runs), 6 teams for Task 2 (a total of 13 runs), and 5 teams for Task 3 (a total of 12 runs).
There are 4 teams successfully submitting technical reports on their methods.
This year, we have 1 first prize, 1 second prize for Task 1; 2 first prize, 1 second prize for Task 2; and 1 first prize, 2 second prize for Task 3.

A variety of methods were used for Task 1: retrieval on only the lexical features, taking advantage of pretrained model, combining the lexical and semantic information, and using boosting technique for classification.
For Task 2, transformer-based \cite{vaswani2017attention} models are prevalent, but other classical machine learning models are also applied. 
There is much room for improvement in this task.
For Task 3, pretrained models are also utilized in the approaches. Participating teams are not provided with a relevant article list for each query.
Interestingly, though, the highest and lowest results in Task 3's rankings are higher than the highest and lowest results in Task 2's rankings.

In COLIEE 2021 on English and Japanese statute law data, pretrained models have an advantage over traditional methods \cite{wehnert2021legal,yoshioka2021bert,nguyen2021paralaw}. This phenomenon is also repeated at ALQAC 2021 on Vietnamese data. This opens up new possibilities in the application of pretrained models in the field of law.
In the coming years, we plan to add more data on legal information retrieval, legal question answering as well as introduce to the community other interesting tasks related to automated legal processing.

\vspace{.5cm}
\section*{Acknowledgments}
This work was supported by JSPS Kakenhi Grant Number 20K20406,  Japan.

\vspace{.5cm}
\bibliographystyle{unsrt}
\bibliography{references.bib}

\end{document}